\documentclass[sigconf, anonymous=false]{acmart}

\usepackage{amsmath,amssymb,amsfonts}
\usepackage{algorithmic}
\usepackage{graphicx}
\usepackage{textcomp}
\usepackage{xcolor}
\usepackage[flushleft]{threeparttable}

\usepackage{makecell}

\usepackage{amsmath,amssymb,amsfonts}
\usepackage{algorithmic}
\usepackage{graphicx}
\usepackage{textcomp}
\usepackage{xcolor}
\usepackage{multirow}
\usepackage{multicol}

\usepackage{threeparttable}
\usepackage{tabularx}
\usepackage{siunitx}
\usepackage{amssymb}
\usepackage{pifont}
\usepackage{bm}

\AtBeginDocument{%
  \providecommand\BibTeX{{%
    \normalfont B\kern-0.5em{\scshape i\kern-0.25em b}\kern-0.8em\TeX}}}

\copyrightyear{2023}
\acmYear{2023}
\setcopyright{acmlicensed}\acmConference[ISWC '23]{Proceedings of the 2023 International Symposium on Wearable Computers}{October 8--12, 2023}{Cancun, Quintana Roo, Mexico}
\acmBooktitle{Proceedings of the 2023 International Symposium on Wearable Computers (ISWC '23), October 8--12, 2023, Cancun, Quintana Roo, Mexico}
\acmPrice{15.00}
\acmDOI{10.1145/3594738.3611369}
\acmISBN{979-8-4007-0199-3/23/10}

\usepackage{fancyhdr}

\AtBeginDocument{%

    \addtolength{\footskip}{2.0\baselineskip}%
    
    \fancyfoot[C]{\small{\textcolor{gray}{\textit{This article has been accepted for publication in the Proceedings of the 2022 ACM International Symposium on Wearable Computers (UbiComp / ISWC 2023). \newline © {ACM} {2023}. This is the author's version of the work. It is posted here for your personal use. Not for redistribution. The definitive Version of Record was published in ACM Library, http://dx.doi.org/10.1145/3594738.3611369.}}}}%
}

\begin{document}

                            


\title{Evaluating Spiking Neural Network On Neuromorphic Platform For Human Activity Recognition
}

\author{Sizhen Bian}
\email{sizhen.bian@pbl.ee.ethz.ch}
\affiliation{%
  \institution{ETH Zürich}
  \city{Zürich}
  \country{Switzerland}
}

\author{Michele Magno}
\email{michele.magno@pbl.ee.ethz.ch}
\affiliation{%
  \institution{ETH Zürich}
  \city{Zürich}
  \country{Switzerland}
}

\renewcommand{\shortauthors}{Sizhen Bian and Michele Magno}

\begin{abstract}
Energy efficiency and low latency are crucial requirements for designing wearable AI-empowered human activity recognition systems, due to the hard constraints of battery operations and closed-loop feedback. While neural network models have been extensively compressed to match the stringent edge requirements, spiking neural networks and event-based sensing are recently emerging as promising solutions to further improve performance due to their inherent energy efficiency and capacity to process spatiotemporal data in very low latency. This work aims to evaluate the effectiveness of spiking neural networks on neuromorphic processors in human activity recognition for wearable applications. The case of workout recognition with wrist-worn wearable motion sensors is used as a study. A multi-threshold delta modulation approach is utilized for encoding the input sensor data into spike trains to move the pipeline into the event-based approach. The spikes trains are then fed to a spiking neural network with direct-event training, and the trained model is deployed on the research neuromorphic platform from Intel, Loihi,  to evaluate energy and latency efficiency. Test results show that the spike-based workouts recognition system can achieve a comparable accuracy (87.5\%) comparable to the popular milliwatt RISC-V bases multi-core processor GAP8 with a traditional neural network ( 88.1\%) while achieving two times better energy-delay product (0.66 \si{\micro\joule\second} vs. 1.32 \si{\micro\joule\second}). 

\end{abstract}




\received{20 February 2007}
\received[revised]{12 March 2009}
\received[accepted]{5 June 2009}

\maketitle

\section{Introduction}

Human activity recognition(HAR) is an active research topic in ubiquitous computing and human-computer interaction that aims to digitalize human behaviors and provide feedback for a better understanding and assistance, especially in healthcare \cite{bhattacharya2022leveraging, zhang2022eatingtrak}, sports science \cite{zhou2016never, hoelzemann2023hang,bian2019passive}, and rehabilitation \cite{long2023dual, halloran2019remote}.
With the increasing popularity of wearable devices, more complex HAR tasks have been moved from the laboratory to real-life scenarios powered by local AI solutions \cite{wang2020fann, bakar2022protean, bian2021capacitive}. However, processing data with machine learning and neural network models on wearables still presents many research challenges due to limited computational resources and power constraints. 
This has led to exploring various neural networks compressions strategies, such as pruning and quantization, to have tiny machine learning models that work with a kilobyte of memory and Giga operations per second in the best case \cite{deng2020model, han2015deep}.

\begin{table*}[!t]
\centering
\caption{SNN explorations in ubiquitous computing}
\label{relatedwork}
\begin{threeparttable}
\begin{tabular}{ p{0.8cm} p{1.8cm}  p{1.2cm} p{1.2cm} p{1.7cm} p{1.2cm} 
 p{1.0cm} p{1.5cm} p{1.8cm} p{1.8cm} }
\hline
Authors/ Year & Application & Sensor/ Dataset  & Neuromor- phic platform &  Encoding & Training    &  Accuracy (vs ANN) &  Inference Latency \tnote{a}\newline(\si{\milli\second })   & Inference Energy \newline (\si{\milli\joule})   & Energy-Delay Product \newline (\si{\micro\joule\second}) \\
\hline
\cite{corradi2019ecg}-2019 & Heartbeat Classification &  ECG\newline MIT-BIH & DYNAP & Delta Modulation & SVM+ rSNN  & 95.6\% (94.2\% )  & NA\tnote{c}  & NA & NA \\
\hline

\cite{sharifshazileh2021electronic}-2021 & Oscillation Detection &  EEG\newline iEEG-HFO & DYNAP-SE & Delta Modulation & Direct SNN  & 78.0\%  (67.0\%)\tnote{d}    & NA  & NA & NA \\
\hline

\cite{ceolini2020hand}-2020 & Hand Gesture Recognition &  EMG \newline Customized & Loihi & Delta Modulation & Direct SNN  & 55.7\%  (68.1\%)    & 5.9 (3.8 on GPU)  & 0.173 (25.5 on GPU) & 1.0 (97.3 on GPU) \\
\hline

\cite{blouw2019benchmarking}-2019 & Key Word Spotting &  Audio \newline Customized & Loihi & Rate Encoding & ANN-to-SNN  & 97.9\%  (97.9\%) \tnote{e}    & 3.38 (1.30 on GPU, 2.4 on Jetson)  & 0.27 (29.8 on GPU, 5.6 on Jetson) & 0.91 (38.7 on GPU, 13.44 on Jetson) \\
\hline

\cite{bos2023sub}-2023 & Ambient Audio Classification &  Audio \newline QUT-NOISE & Xylo & Power band bin to spike & Direct SNN & 98.0\%  (97.9\%)    &  100\tnote{f}  & 0.0093 \tnote{g}(0.25: MAX78000, 11.2: Cortex) & 0.93 \\
\hline

\textbf{Ours} &  \textbf{Human Activity Recognition} &  \textbf{IMU, Capacitive \newline RecGym} & \textbf{Loihi} & \textbf{MT Delta Modulation} & \textbf{Direct SNN} & \textbf{87.5\%  (88.1\%) }   & 4.4 (3.2 on GAP8)  & 0.15 (0.41 on GAP8) & 0.66 (1.31 on GAP8) \\
\hline
\end{tabular}
\begin{tablenotes}
\setlength{\columnsep}{0.8cm}
\setlength{\multicolsep}{0cm}
  \begin{multicols}{2}
            \item[a] Time elapsed between the end of the input and the classification.
            \item[b] Only dynamic energy is considered in Loihi.
            \item[c] Not Available.
            \item[d] HFO was detected with morphology detector \cite{fedele2017resection}.
            \item[e] True Positive.
            \item[f] Median classification latency (from the onset of an audio sample until the first spike from the correct class output neuron).
            \item[g] Dynamic energy consumption.
    \end{multicols}
\end{tablenotes}  
\end{threeparttable}
\end{table*}

While exploring the energy and latency performance of the edge tiny neural networks, the spiking neural network (SNN), a network that mimics the biological neurons in the brain, is also emerging as a promising technology for energy-efficient edge computing. The increase in interest is due to the event-driven nature and its ability to process spatio-temporal data in real-time with low latency and energy efficiency. 
In traditional neural networks, the activation function is typically differentiable, allowing the backpropagation to train a network. However, in SNNs, the firing of a neuron is a discrete event rather than a differentiable continuous function; thus, traditional backpropagation cannot be used to train SNNs. Two approaches have been explored in the past years to overcome this challenge, direct SNN training with surrogate gradient \cite{neftci2019surrogate, ledinauskas2020training} and ANN-to-SNN by first accumulating the events \cite{gehrig2019end, wu2021training, li2022wearable, fra2022human}. While falling behind in reaching state-of-the-art accuracy, direct training is more biologically similar and preserves the temporal resolution of the spikes.
To fully make use of the biological plausibility and validate the envelope of energy and latency efficiency of SNN in HAR, this work will use the surrogate gradient approach for direct event training of a workout recognition dataset, as a case study of neuromorphic solution for HAR. The neuromorphic chip Loihi will be used to profile the SNN performance
taking the state-of-the-art performance from the general ANN on advanced edge processors with dedicated hardware accelerators as the baseline.


\section{Related work and contribution}

The previous application works with SNN are mostly focused on vision tasks with dynamic vision sensors (DVS) \cite{li2019132, akrarai2020novel}, where the input data is a sequence of visual events. 
A few edge explorations with DVS were also presented to validate the performance of end-to-end neuromorphic platforms from sensing to computing  \cite{rutishauser2023colibries}. Compared with vision applications, exploring SNN on low-dimensional data like audio and sensor signals is relatively new and much less explored \cite{davies2021advancing}. However, recent studies have shown promising results in using SNNs for ubiquitous computing with low-dimensional signal sensors, which can provide essential insights into related topics like HAR. Table \ref{relatedwork} lists several recent studies that explore SNNs for ubiquitous computing with low dimensional signals and their resulting performance in different applications. Kyle et al. \cite{buettner2021heartbeat} and Federico et al. \cite{corradi2019ecg} explored the heartbeat classification with SNN with different training strategies and validated it on two neuromorphic processors with competitive accuracy. The energy-delay product (EDP) on Loihi shows over twenty-eight times more efficiency than the inference on a CPU. 
Enea et al. \cite{ceolini2020hand} run a direct-trained SNN on Loihi with customized EMG and DVS data set for hand gesture recognition. Similar to \cite{buettner2021heartbeat}, the EMG results on Loihi outperform in EDP compared with the results on GPU by ninety-seven times more efficiently. Besides the biological signals, audio signals were also explored with SNN \cite{blouw2019benchmarking}.
In \cite{bos2023sub}, a fresh edge neuromorphic processor, Xylo, was used to classify ambient audios. An impressive inference energy was reported on Xylo with only 9.3\si{\micro\joule}, over twenty-six times less energy than the edge IoT processor MAX78000 owning a convolutional hardware accelerator. One common result of those SNNs on low-dimensional signals is that SNN supplies state-of-the-art inference energy and impressive EDP compared with ANN on CPU and GPU. Besides this, the ANN-to-SNN training approach often results in competitive accuracy while the direct trained SNN shows incompetence in accuracy compared with the ANN result\cite{ceolini2020hand}. The reason is that while encoding the signal to spikes, for example, the delta modulation, information loss is happening, especially for fast and huge signal variations \cite{ceolini2020hand, vitale2022neuromorphic, auge2021survey}. 

In this work, we bring the following contributions:

\begin{enumerate}
\item We demonstrated the feasibility of using SNN for sensor-based HAR tasks pursuing latency and energy efficiency with a direct-trained SNN on the neuromorphic platform Loihi. The first Spiking-IMU dataset and the corresponding direct-trained SNN are released for benchmarking of HAR with the neuromorphic solution \footnote{https://github.com/zhaxidele/HAR-with-SNN}.
\item With spike trains generated by a multi-threshold delta modulation approach, a comparable accuracy (87.5\%) is achieved compared with the ANN approach on the novel IoT processor GAP8 (88.1\%), which has a dedicated RISC-V cluster for hardware acceleration and presented the state of the art edge AI performance in a rich of applications.
\item The latency and energy efficiency of the neuromorphic approach HAR and the mainstream approach HAR were compared in this case study, and it showed that the neuromorphic approach of HAR using SNNs on Loihi outperforms the ANN method in terms of inference energy on GAP8 while falling behind lightly in latency. However, the neuromorphic approach shows nearly two times the energy-delay product (0.66 \si{\micro\joule\second} vs. 1.31 \si{\micro\joule\second}).

\end{enumerate}

\begin{figure*}[!t]
\centering
  \includegraphics[width=0.85\textwidth, height=3.5cm]{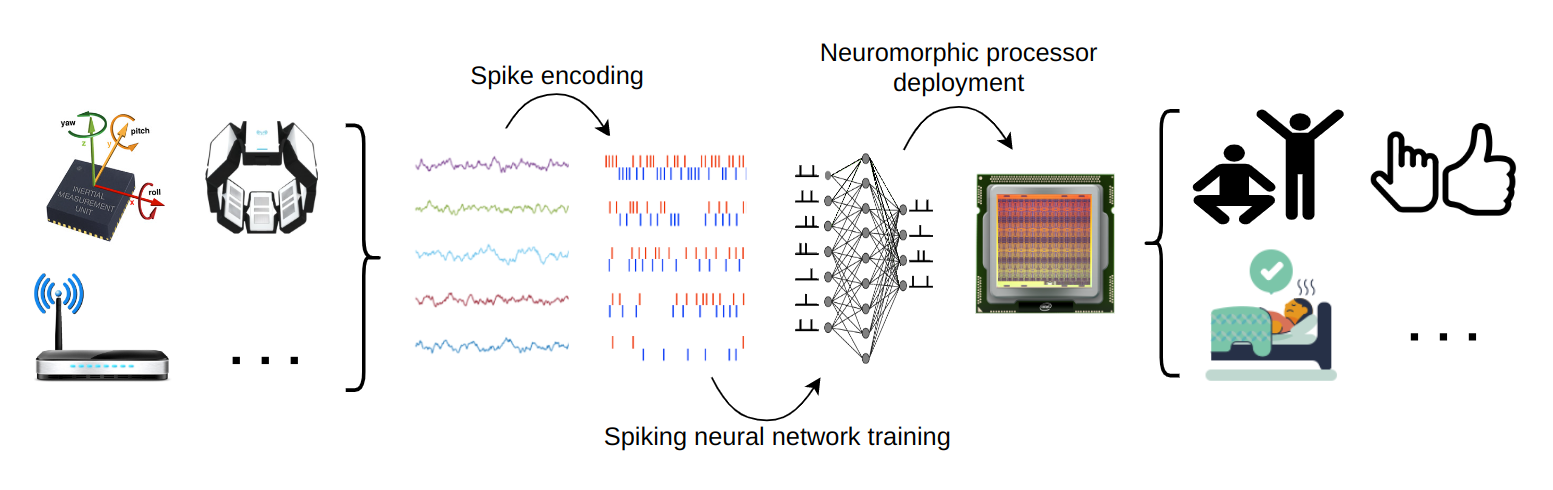}
  \caption{Human activity recognition using the spiking neural network where the network is processed on neuromorphic platforms pursuing energy and latency efficiency}
  \label{fig:teaser}
\end{figure*}

\section{System Architecture}

Figure \ref{fig:teaser} depicts the pipeline of the proposed SNN for HAR applications, including three key steps: spike encoding from sensor data, off-line SNN training, and on line SNN inference on neuromorphic processor. To have a fair comparison with neural networks and low power digital processors, in this work we use a public data set, RecGym \cite{bian2022contribution}, as a case study. However, it is important to notice that the proposed approach can be used with other HAR-related data sets from various sensing modalities. The data set records ten volunteers' gym sessions with a sensing unit composed of an IMU sensor and a Body Capacitance sensor\cite{bian2022using, bian2021systematic}. The sensing units were worn at three positions: on the wrist, pocket, and calf. Twelve gym activities are recorded, including eleven workouts like ArmCurl, LegPress, and StairsClimber, and a "Null" activity when the volunteer hangs around between different workout sessions. Each participant performed the selected workouts for five sessions in five days. Altogether, fifty sessions of gym workout data are presented in this data set. In this study, we only focus on the motion signals with the sensing unit worn on the wrist.

\subsection{Spike encoding}

\begin{figure}[!t]
\begin{minipage}[t]{0.79\linewidth}
\centering
\includegraphics[width=0.99\textwidth,height=3.5cm]{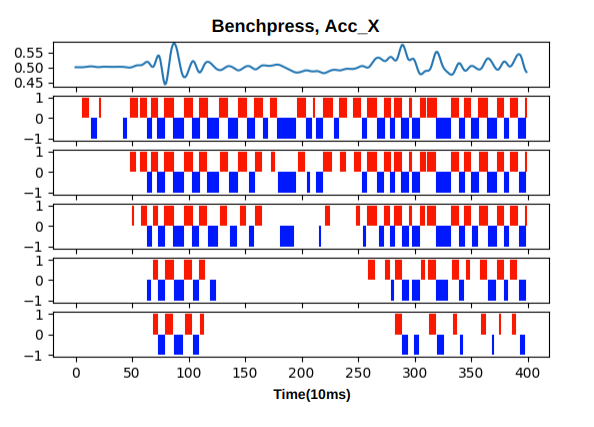}
\caption{Encoded spike trains of signal Acc\_x from the workout of BenchPress with thresholds of [0.00005, 0.0001, 0.0002, 0.0004, 0.0008]}
\label{Spiketrain}
\end{minipage}
\end{figure}

To directly train an SNN, traditional numerical sequential value needs to be transformed into spike streams that carry both temporal and spatial knowledge of the original signals. Different encoding approaches have been explored mostly for vision data transformation \cite{guo2021neural}, like latency encoding, rate encoding, delta modulation, etc. Each has advantages and limitations and has been adopted in different works  \cite{ auge2021survey}. For example, latency encoding normally achieves the best processing latency and energy consumption with fewer synaptic operations while being more susceptible to noise. Rate coding is demonstrated to exist in sensory systems like the visual cortex and motor cortex \cite{srivastava2017motor}, showing the best resilience to input noise while limited by a lengthy processing period. In this work, we used the delta modulation approach due to the optimal trade of complexity and latency from both the firmware and hardware implementation.  Moreover, the analog sensory information can be directly encoded to the spike train at the front end. To address the accuracy degradation caused by information loss during the encoding, we gave multiple thresholds for spike train generation. The relationship between the continuous signal $s(t)$
and its spiking counterpart $\hat{s}(t)$ is given by Equation \ref{eq: relationship}.
\begin{equation}
  \setlength{\arraycolsep}{0pt}
  \hat{s_i}(t) = \left\{ \begin{array}{ l l }
      1, \text{ if }s(t) - s(t-1) >  \epsilon_i \\
      -1, \text{ if }s(t) - s(t-1) < -\epsilon_i
    \end{array} \right. \\
  \label{eq: relationship}
\end{equation}
where $\epsilon_i$ is the threshold empirically chosen for spike encoding, and $i$ $(0\sim4)$represents the index of applied thresholds list. Figure \ref{Spiketrain} depicts the five spike train channels encoded from the X-axis of the accelerometer, where the fast and large signal trend gives spikes in more spike train channels. For the inertial data, the thresholds were empirically set to 0.00005 x $(i+1)$, while 0.0000125  x $(i+1)$ for the capacitance data, and the threshold element number was empirically set to five. In future work, a systematical exploration of choosing the best threshold value and element number will be explored. As seven continuous signals were collected in the data set, we got thirty-five (7x5) spike trains for SNN training and inferring. With two seconds time window as a classification instance, we got 81291 spiking samples for building the SNN model.

\subsection{Spiking neural network and Loihi}

One of main contribution of this paper is the design of a SNN model and its evaluation on the Intel Loihi research platform. 
Loihi \cite{davies2018loihi} is an asynchronous neuromorphic digital processor mainly for research. The processor consists of a many-core mesh of 128 neuromorphic cores for spike processing and three synchronous Lakemont x86 cores to monitor and configure the network and assist with the injection and recording of input/output spikes. Each neuro core in Loihi can access its local memories independently without needing to share a global memory bus and can implement up to 1024 current-based leaky integrate and fire neurons. Among other research platforms, Loihi has been selected because it includes a software SDK to design and profile proposed SNN. 

Our proposed SNN is composed of two convolutional and two dense layers (32C64C128D12D) with a kernel size of three, as Table \ref{model} lists. The threshold for neuron spiking was empirically selected. The current and voltage decay constants for the leaky integrate and fire neurons were set to 1024 (32 ms) and 128 (4 ms), respectively. Before spike encoding, the data set was interpolated to 1 kHz using the univariate spline method, aiming to maximally approach the biological behaviors of the brain regarding information feeding. Each sample contains two seconds length of spike trains. The model was trained offline on GPU with weighted classes and leave-one-user-out cross-validation, and the trained weights and delays were then used to configure the network on Loihi hardware for inference purposes.

To fully make use of the biological plausibility of SNN, we used the framework SLAYER \cite{shrestha2018slayer} for direct training, aiming to pursue the envelope of energy and latency efficiency of SNN. SLAYER evaluates the gradient of the convolutional and linear layers in an SNN by a temporal credit assignment policy, which distributes the error credit of an error back both through layers and in time, as a spiking neuron’s current state depends on its previous states. Then a probability density function is used to estimate the change in the neuron state, thus approximating the derivative of the spike function. With SLAYER, the synaptic weights and axonal delays can be trained, and some state-of-the-art performances have been achieved on neuromorphic datasets like the NMNIST and ibmDVSGesture \cite{shrestha2018slayer}. 
SLAYER supports Loihi-specific implementation for neuron model and weight quantization.

\begin{table}[htb]
      \caption{SNN model for the spiking RecGym dataset}
      \label{model}
    \begin{threeparttable}
      \begin{tabularx}{\textwidth}{r|r|r|r|r|r}
        & Type & Size & Feature Size & Features & Stride 
        \\
        \cline{1-6}
        0 & Input & 7x5x2 & - & - & - 
        \\
        1 & Conv & 7x5x32 & 3x3 & 32 & 1 
        \\
        2 & Conv & 7x5x64 & 3x3 & 64 & 1 
        \\
        3 & Dense & 2240 & - & 128 & - 
        \\
        4 & Dense & 128 & - & 12 & - 
        \\
        \end{tabularx}
      \end{threeparttable}

\end{table}

\section{Experimental Evaluation}

\begin{table}[!t]
\centering
\caption{Classification profiling vs. general edge solutions}
\label{profiling}
\begin{threeparttable}
\begin{tabular}{ p{1.6cm} p{0.5cm}  p{0.9cm} p{0.8cm} p{0.8cm} p{1.8cm} }
\hline
Hardware & Model & Accuracy  & Latency (\si{\milli\second }) &  Energy (\si{\milli\joule})  & Energy-Delay Product (\si{\micro\joule\second}) \\
\hline
Loihi (Neuromorphic)& SNN  & 87.5\%  & 4.4 & 0.15  & 0.66 \\
\hline
GAP8 \newline(RISC-V) & ANN & 88.1\%   & 3.2 & 0.41 & 1.31 \\
\hline
STM32 (Cortex-M7) & ANN & 89.3\%   & 20.88 & 8.07  & 168.5 \\

\hline

\hline
\end{tabular}
\end{threeparttable}
\end{table}

Table \ref{profiling} lists the workouts classification performance with the trained SNN on Loihi. In comparison, we selected an ANN model using the same data set and being deployed on two different IoT processors, presented in \cite{bian2022exploring}. Such a comparison has seldom been made, as previous SNN evaluations mostly used ANN deployed on GPU/CPU as the baselines. The result will be meaningful for developing ubiquitous neuromorphic edge computing by supplying a straightforward comparison with the state-of-the-art using mainstream solutions. The multi-threshold spike encoding approach results in an accuracy of 87.5\% with the directly trained SNN, which is much better than the single-threshold encoding result (below 60\%) and acceptable compared with the accuracy from the ANN approach considering that the accuracy of direct-trained SNN on spike streams degrades in most cases. The inference latency of SNN on Loihi implies the time elapsed between the end of the input and the classification output and is reported as 4.4 \si{\milli\second }, which is also much better than the latency on general IoT processors like STM32 with Cortex-M7 core but falls behind slightly to the GAP8, which features 8 RISC-V cores for dedicated hardware acceleration. However, the neuromorphic pipeline outperforms in dynamic energy consumption (0.15 \si{\milli\joule}), benefitting from the sparsity of the spike trains and the in-memory computing of Loihi, which results in an EDP of 0.66 \si{\micro\joule\second}, while the EDP on GAP8 and STM32 are almost two times and over two hundred times higher, respectively. The energy reported here is the dynamic energy on Loihi, which is measured by enabling the energy probe during inference as the difference between the total energy consumed by the network and the static energy when the chip is idle. We have to acknowledge that Loihi is not for edge computing specifically. Instead, it is designed more for general-purpose neuromorphic research. Thus there is still space for raising the neuromorphic performance, for example, the spike injection speed (the primary x86 core always waits 1ms before allowing the system to continue to the next timestep). To have a more fair comparison, end-to-end solutions of neuromorphic and traditional approaches should be developed, adopting the newly released edge neuromorphic processors \cite{bos2023sub, rutishauser2023colibries}.

\section{Conclusion}

This work explored the neuromorphic solution of human activity recognition with a typical case study of workout recognition. Neuromorphic solutions, mainly inferring the SNN associated with the neuromorphic processor, have been emerging benefiting from its latency and energy efficiency. We started with a multi-threshold delta modulation to encode the raw motion sensor signal into multiple spike trains, aiming to reduce the information loss during spike generation. A shallow SNN model was then used to train the spike-form workouts signal with the SLAYER framework. The model runs on Loihi showed a comparable accuracy of  87.5\% and an impressive energy-delay-product of 0.66 \si{\micro\joule\second}, compared with the state-of-the-art ANN solution on GAP8. This work demonstrates the efficiency of neuromorphic solutions in ubiquitous computing that pursues latency and energy efficiency. For future work, we will focus on new features in neuromorphic solutions that exceed the traditional edge solutions, for example, learn on the fly that can adapt the SNN models for specific subjects and environments, boosting the inference accuracy. We will also explore the newly released edge neuromorphic platforms and Loihi2, which has redesigned asynchronous circuits supplying faster speed and enhanced learning capabilities, featuring multiple times performance boosting compared with its predecessor.

\begin{acks}
This work was supported by the CHIST-ERA project ReHab(20CH21-203783).
\end{acks}

\bibliographystyle{ACM-Reference-Format}
\bibliography{sample-base}

\end{document}